\documentclass[lettersize,journal]{IEEEtran}
\usepackage{cite}
\usepackage{geometry}
\usepackage{amsmath,amssymb,amsfonts}
\usepackage{graphicx}
\usepackage{textcomp}
\usepackage{xcolor}
\def\BibTeX{{\rm B\kern-.05em{\sc i\kern-.025em b}\kern-.08em
    T\kern-.1667em\lower.7ex\hbox{E}\kern-.125emX}}
\usepackage{makecell}
\usepackage{booktabs}
\usepackage{multirow}
\usepackage{bbding}
\usepackage{subfigure}
\usepackage{array}
\usepackage[caption=false,font=normalsize,labelfont=sf,textfont=sf]{subfig}
\usepackage{textcomp}
\usepackage{stfloats}
\usepackage{url}
\usepackage{verbatim}
\usepackage{balance}

\usepackage{amsmath,amsfonts}
\usepackage{algorithm}
\usepackage{algpseudocode}
\begin{document}

\title{A CT-guided Control Framework of a Robotic Flexible Endoscope for the Diagnosis of the Maxillary Sinusitis\\}
%{\footnotesize \textsuperscript{*}Note: Sub-titles are not captured in Xplore and
%should not be used}

\author{Puchen Zhu, Huayu Zhang, Xin Ma*, Xiaoyin Zheng, Xuchen Wang, Kwok Wai Samuel Au}

\maketitle

\begin{abstract}
Flexible endoscopes are commonly adopted in narrow and confined anatomical cavities due to their higher reachability and dexterity. However, prolonged and unintuitive manipulation of these endoscopes leads to an increased workload on surgeons and risks of collision. To address these challenges, this paper proposes a CT-guided control framework for the diagnosis of maxillary sinusitis by using a robotic flexible endoscope. In the CT-guided control framework, a feasible path to the target position in the maxillary sinus cavity for the robotic flexible endoscope is designed. Besides, an optimal control scheme is proposed to autonomously control the robotic flexible endoscope to follow the feasible path. This greatly improves the efficiency and reduces the workload for surgeons. Several experiments were conducted based on a widely utilized sinus phantom, and the results showed that the robotic flexible endoscope can accurately and autonomously follow the feasible path and reach the target position in the maxillary sinus cavity. The results also verified the feasibility of the CT-guided control framework, which contributes an effective approach to early diagnosis of sinusitis in the future.

\end{abstract}


\section{Introduction}
Sinusitis has long been a prevalent respiratory inflammatory, affecting millions of people worldwide with an estimated incidence of 10–15\%\cite{ref1}. 
Without early diagnosis and treatment, patient conditions may deteriorate.
However, the early diagnosis of maxillary sinusitis (DMS) continues to present numerous challenges \cite{ref2}.
Conventional DMS for early diagnosis requires surgeons to manipulate a medical endoscope during the diagnosis to reach the target position in the maxillary sinus cavity.
Although these medical endoscopes are intuitive for surgeons to manipulate, rigid distal ends and limited dexterity make it hard to reach the narrow and confined space of the maxillary sinus \cite{ref3}.
Moreover, prolonged manipulation of these medical endoscopes during the diagnosis greatly increases the workload of surgeons. 

\begin{figure}[!t]
\includegraphics[width=0.5\textwidth]{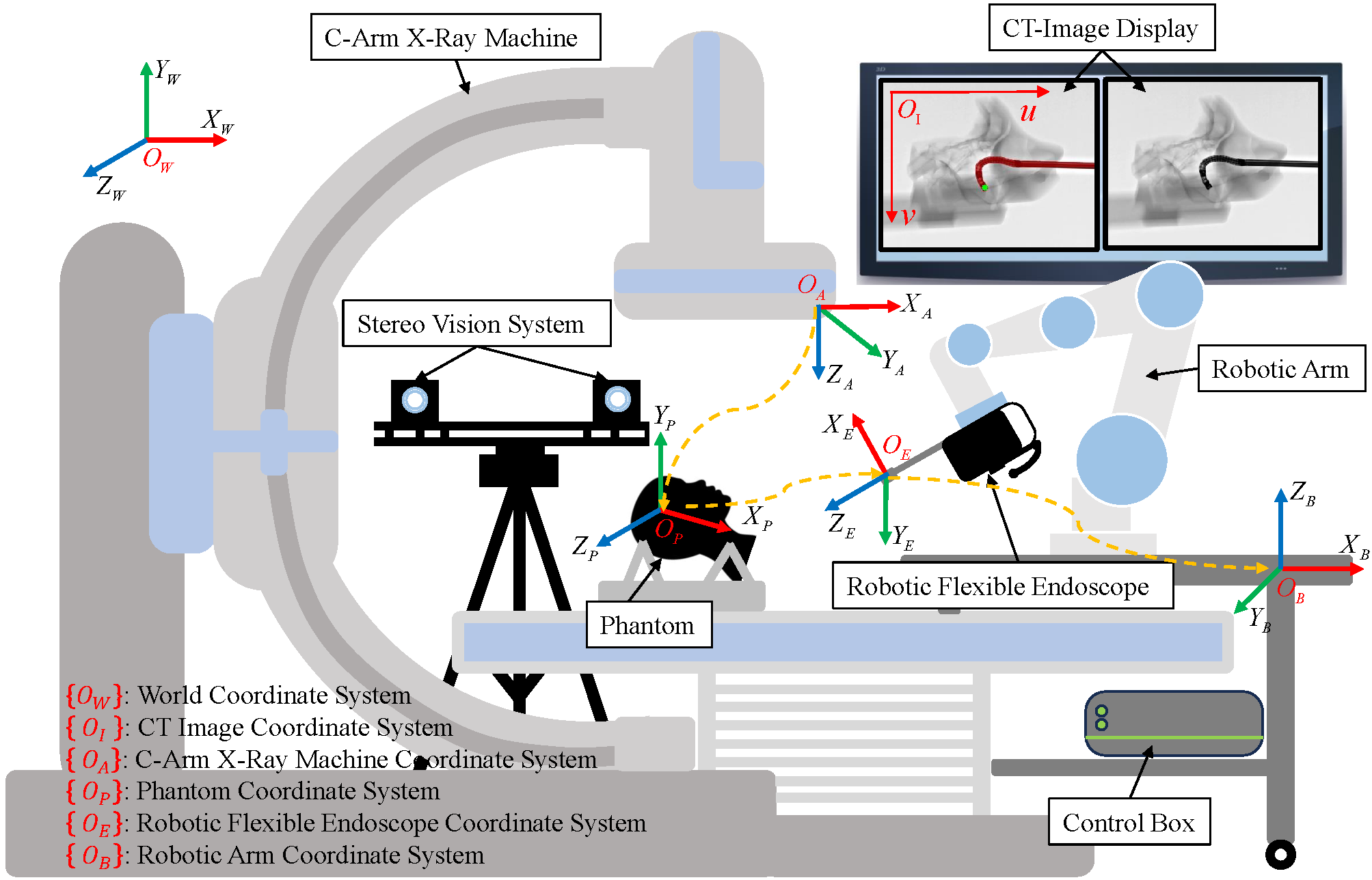}
\centering
\caption{Overview of the CT-guided robotic flexible endoscope for the diagnosis of the maxillary sinusitis}
\label{Overview}
\end{figure}

To address the aforementioned challenges, lots of flexible endoscopes are employed \cite{ref4,ref5,instru}.
Compared with widely used rigid endoscopes, flexible endoscopes provide greater dexterity at the distal end allowing it to reach the maxillary sinus cavity for early diagnosis, which is verified in many clinical researches\cite{ref6,ref7}. 
In addition, many state-of-the-art robot-assisted and hybrid-structure robotic endoscopes are developed, providing significant assistance to surgeons during the DMS. 
Hong \textit{et al.} \cite{ref8} proposed a two-segment tendon-driven flexible robotic endoscope that can accurately reach the target space in the maxillary sinus model, which is demonstrated by the experiments. 
A dual master-slave robotic system including a flexible endoscope is developed by Yoon \textit{et al.} \cite{ref9}, in which surgeons can easily manipulate the endoscope with a master device. 
Wang \textit{et al.} \cite{ref10} presented a hybrid-structure hand-held robotic endoscope, which is well-designed to offer sufficient dexterity. This endoscope can also be installed on a robotic platform to reduce surgeon's fatigue.

Compared with widely used traditional rigid endoscopes, flexible endoscopes have higher dexterity while surgeons cannot intuitively manipulate them for DMS.
Therefore, many control methods have been employed for flexible endoscopes. Initially, some hand-held flexible endoscopes are manually controlled \cite{ref10,ref11,ref12}. Subsequently, more efficient and intuitive teleoperation control methods are performed by using master devices \cite{ref13,ref14,ref15}. In addition, Sivananthan \textit{et al.} \cite{ref16} proposed a novel gaze-controlled flexible robotized endoscope, which is reported to be useful and satisfying. In order to facilitate surgeons to control both endoscope and instrument, several teleoperated robotic systems with foot control for flexible endoscopic surgery are designed and performed \cite{ref17,ref18}. Zuo \textit{et al.} \cite{ref19} developed a wearable hands-free human-robot interface to control all the DOFs of a flexible endoscope at the same time. 

Nevertheless, the absence of closed-loop feedback in the aforementioned control methods leads to diminished positional accuracy at the distal end of flexible endoscopes. Consequently, such limitations increase the potential for collisions between the endoscopes and tissues or polyps during DMS procedures. Furthermore, the flexible endoscope is unable to realize autonomous navigation by using these control methods, and the surgeons still have to experience fatigue. 
Currently, there is an increasing tendency for flexible endoscopes to implement image-guided control methods \cite{ref20}. 
Deng \textit{et al.} \cite{ref21} proposed a novel image-guided robotic nasotracheal intubation system, which achieves automatic control of the endoscope. 
Lee \textit{et al.} \cite{ref22} presented an image-guided endoscopic sinus surgery system, which is demonstrated to have reliable potential for clinical application according to the clinical study.  
However, these image-guided approaches ignore that issues and polyps may cover the lens of the endoscope, which makes the controller lose image information of the target position leading to a wrong decision.
The other idea to effectively utilize the image information is by using the Computerized Tomography (CT) scan \cite{ref23}, but most studies just employed the CT scanning to conduct evaluation before the sinus surgery. This means that the CT technology cannot be regarded as a guide to provide image information as feedback during DMS.

Thus, to address the aforementioned issues, this paper achieves a fusion of CT technology and a robotic flexible endoscope, as shown in Fig. \ref{Overview}, and a novel CT-guided control framework of the robotic flexible endoscope is proposed for DMS. 
When conducting the CT-guided control framework on a realistic 3D sinus phantom, the robotic flexible endoscope is able to autonomously and accurately follow the feasible path and reach the target position in the maxillary sinus cavity. 
The contributions of this paper are as follows:
\begin{itemize}
\item Proposal of a novel CT-guided control framework, which is the first time to autonomously control the robotic flexible endoscope to efficiently and accurately follow a feasible path and reach the target position in the maxillary sinus cavity.
\end{itemize}
\begin{itemize}
\item Demonstration of the efficiency, accuracy, and feasibility of the CT-guided control framework based on a realistic 3D sinus phantom.
\end{itemize}

The rest of this paper is organized as follows. Section II introduces the overall control framework, including the overview of the framework, path planning for the flexible endoscope, systems registration, real-time image process, and optimal control scheme. In section III, several experiments are conducted and the experimental results are shown in detail. Section IV is the conclusion of this paper.

\section{Overall Control Framework}
In this section, a CT-guided control framework of the robotic flexible endoscope is proposed for the diagnosis of maxillary sinusitis. 
The CT-guided control framework is divided into five parts, which are successively described in detail.

\subsection{Overview of the Control Framework}
\begin{figure*}[!t]
\includegraphics[width=1\textwidth]{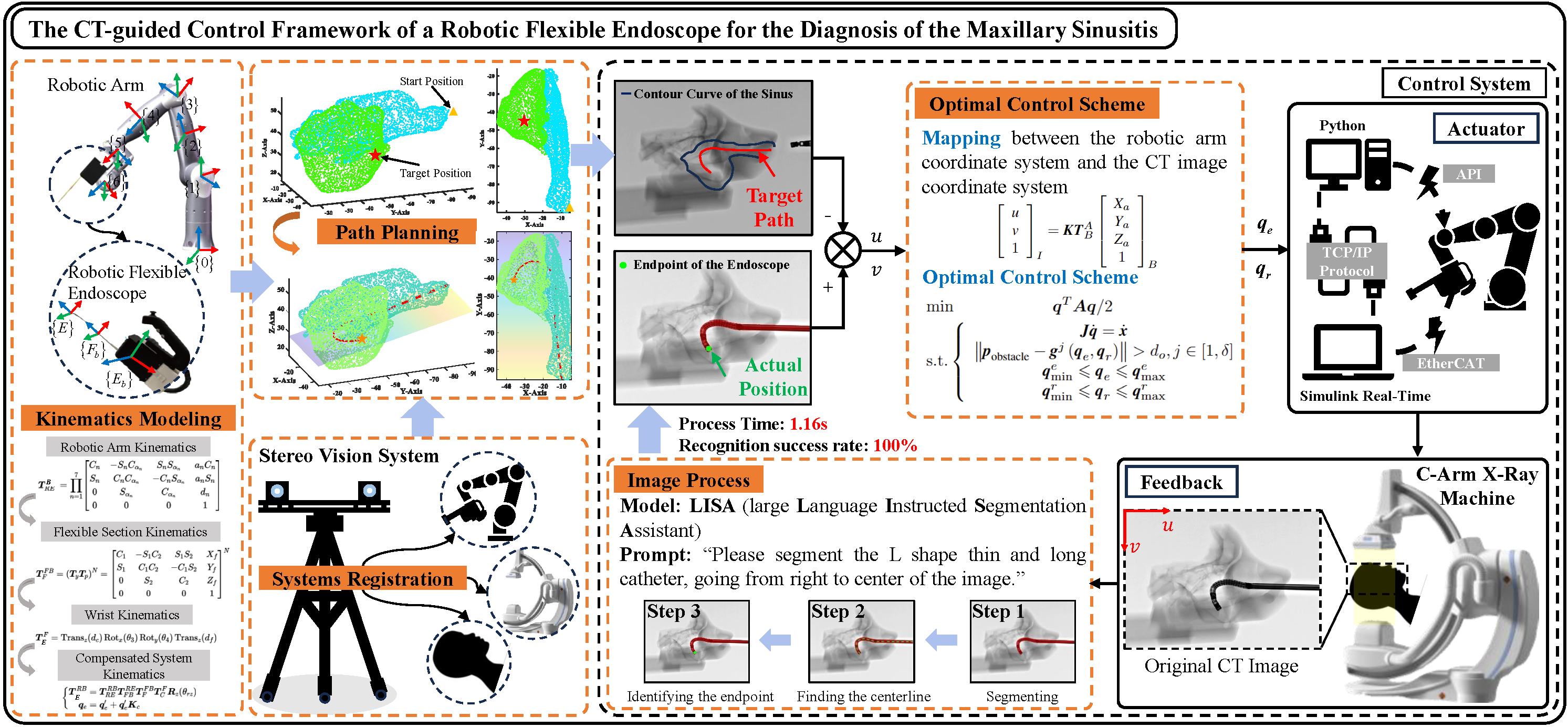}
\centering
\caption{Overview of the CT-guided control framework of the robotic flexible endoscope for diagnosis of maxillary sinusitis, including five modules: 1) kinematics modeling module, 2) path planning module, 3) systems registration module, 4) image process module, and 5) control system module.}
\label{framework}
\end{figure*}

As shown in Fig. \ref{framework}, 
the kinematics modeling of the system includes the kinematics modeling of the robotic flexible endoscope and that of the robotic arm. Then, the pose relationship between the endpoint of the endoscope and the base of the robotic arm is constructed. 
To implement path planning for the robotic flexible endoscope, a feasible path is determined via a common 3D path planning algorithm based on cloud points of the maxillary sinus from CT scan data. 
The robotic flexible endoscope can efficiently and accurately follow the path to reach the target position in the maxillary sinus cavity with improved visual coverage \cite{ref10}, and to avoid obstacles. 
In systems registration, a stereo vision system is used to realize the registration among the robotic arm coordinate system, the robotic flexible endoscope coordinate system, the C-arm X-ray machine coordinate system, and the phantom coordinate system. It is noted that all coordinate systems in the framework are unified on the robotic arm coordinate system. 
The acquisition and annotation of CT images are time-consuming and costly. Due to the poor robustness of traditional algorithms, a multi-modality large language model is employed to accurately identify the endoscope's endpoint position in the image process system. 
In the design of the control system, an optimal control scheme is proposed. The difference between the target and actual position in the robotic arm coordinate system of the endpoint of the robotic flexible endoscope serves as feedback information, enabling the control system to form a closed loop. The closed-loop feedback control system ensures the accurate and efficient control of the CT-guided robotic flexible endoscope.

%Due to the acquisition and annotation of CT images is time intensive and costly, and the traditional algorithm has poor robustness, the multi-modality large language model is used to identify the end point position of the endoscope. 
\subsection{Kinematics Modeling of the System}
The robotic flexible endoscope consists of a 2-DOF flexible section that enables the flexible endoscope to reach the target position in the maxillary sinus cavity, and a 2-DOF rigid wrist that leads to improved visual coverage. 
The kinematics modeling of the system is described as the mapping between image space and task space, and task space and configuration space. The configuration space consists of robotic arm configuration space $\textbf{\textit{q}}_r=\left[\theta_1^r, \ldots, \theta_7^r\right]$ and robotic flexible endoscope configuration space $\textbf{\textit{q}}_e=\left[\theta_1^e, \theta_2^e, \theta_3^e, \theta_4^e\right]$. 
In addition, the mappings between motor space and tendon space, and tendon space and configuration space are introduced in our previous work \cite{ref10}. 
%In order to reach the maxillary sinus cavity, the robotic arm is required to provide axial displacement of the endoscope, which assists the robotic flexible endoscope in navigating in the nasal cavity. 
%When approaching the maxillary sinus ostium, the robotic flexible endoscope is required to bend through the ostium and access the maxillary sinus cavity. 
%During this coupling motion, the mapping between image space and task space can be described as:
%\begin{equation}
%\textbf{\textit{K}}\left[\begin{array}{cc}
%\Delta u & 0 \\
%0 & \Delta v
%\end{array}\right] \textbf{\textit{K}}^T=k %\textbf{\textit{I}}\left[\begin{array}{cc}
%\textbf{\textit{R}} & \textbf{\textit{P}} \\
%0 & 1
%\end{array}\right]=k \textbf{\textit{I}} \textbf{\textit{T}}_C^{R %B}
%\end{equation}
%where \textbf{\textit{K}} is a 4-by-2 constant coefficient matrix, each element in \textbf{\textit{K}} is off-line calibrated, $\textbf{\textit{K}}^T$ is the transpose of \textbf{\textit{K}}, $\Delta u$ and $\Delta v$ represent the differences in the $u$ and $v$ directions between the current image coordinates and the expected image coordinates of the endpoint of the endoscope, $k$ is scale factor, \textbf{\textit{I}} is a 4-by-4 unit matrix. \textbf{\textit{P}} is a 3-by-1 translation matrix, representing the 3D position of the CT-guided robotic flexible endoscope in task space, \textbf{\textit{R}} represents orientation in task space. $\textbf{\textit{T}}_E^{F B}$ describes the transformation between the endpoint of the endoscope and the base of the robotic arm.
By using the Denavit-Hartenberg (D-H) method, the transformation $\textbf{\textit{T}}_{R E}^{B}$ from the end-effector of the robotic arm to the base of the robotic arm is derived as follows.
\begin{equation}
\textbf{\textit{T}}_{R E}^{B}=\prod_{n=1}^7\left[\begin{array}{cccc}
C_n & -S_n C_{\alpha_n} & S_n S_{\alpha_n} & a_n C_n \\
S_n & C_n C_{\alpha_n} & -C_n S_{\alpha_n} & a_n S_n \\
0 & S_{\alpha_n} & C_{\alpha_n} & d_n \\
0 & 0 & 0 & 1
\end{array}\right]
\end{equation}
where $C_n$ and $S_n$ represent $\cos \left(\theta_n^r\right)$ and $\sin \left(\theta_n^r\right)$, respectively. $C_{\alpha_n}$ and $S_{\alpha_n}$ represent $\cos \left(\alpha_n\right)$ and $\sin \left(\alpha_n\right)$, respectively. $\alpha_n$, $d_n$, and $a_n$ are structure parameters of the robotic arm, as listed in the D-H table shown in Table I. The transformation from the distal end of the flexible section to the base of the flexible section $\textbf{\textit{T}}_F^{F B}$ is descrbed as:

\begin{table}[]
\centering
\setlength\tabcolsep{12pt}
\caption{D-H Parameters of the robotic arm}
\begin{tabular}{ccccc}
\toprule
Joint & $a_i$ (mm)   & $\alpha_i$ ($^\circ$)  & $d_i$ (mm)     & $\theta_i$ ($^\circ$)  \\ \midrule
1     & 0    & 0     & 345  & $\theta_1$     \\
2     & 0    & 90  & 65  & $\theta_2$     \\
3     & 0    & -90 & 395  & $\theta_3$+180    \\
4     & 20 & -90 & -55 & $\theta_4$+180    \\
5     & 20 & 90  & 385  & $\theta_5$+180    \\
6     & 0    & 90  & 100    & $\theta_6$+90  \\
7     & 110 & 90  & 55  & $\theta_7$     \\ \bottomrule
\end{tabular}
\end{table}

\begin{equation}
\textbf{\textit{T}}_F^{F B}=\left(\textbf{\textit{T}}_y \textbf{\textit{T}}_p\right)^N=\left[\begin{array}{cccc}
C_1 & -S_1 C_2 & S_1 S_2 & X_f \\
S_1 & C_1 C_2 & -C_1 S_2 & Y_f \\
0 & S_2 & C_2 & Z_f \\
0 & 0 & 0 & 1
\end{array}\right]^N
\end{equation}
where $N$ denotes the pair numbers of orthogonally arranged
notches distributed on the flexible section, $\textbf{\textit{T}}_y$ and $\textbf{\textit{T}}_p$ are the transformations matrices of the 2-DOF motion (yaw and pitch) of the flexible section. $C_1$ and $C_2$ represent $\cos \left(\theta_1^e\right)$ and $\cos \left(\theta_2^e\right)$, respectively. $S_1$ and $S_2$ represent $\sin \left(\theta_1^e\right)$ and $\sin \left(\theta_2^e\right)$, respectively. $\left[X_f, Y_f, Z_f\right]^T$ denotes the 3D position of the distal end of the flexible section. The transformation from the distal end of the flexible section to the endpoint of the endoscope $\textbf{\textit{T}}_E^F$ is also derived as follows.
\begin{equation}
\textbf{\textit{T}}_E^F=\operatorname{Trans}_z\left(d_{c}\right) \operatorname{Rot}_x\left(\theta_3\right) \operatorname{Rot}_y\left(\theta_4\right) \operatorname{Trans}_z\left(d_{f}\right)
\end{equation}
where $\operatorname{Trans}_z(\cdot)$, $\operatorname{Rot}_x(\cdot)$, and $\operatorname{Rot}_y(\cdot)$ are the homogeneous transformation matrices for the translation and rotation about the $z$,
$x$, and $y$ axes, respectively. $d_{c}$ and $d_{f}$ are lengths of two rigid joints. However, the kinematic model of the robotic flexible endoscope has limited accuracy due to errors from manual assembly and hysteresis. Therefore, several empirical kinematic compensation terms are considered in robotic flexible endoscope kinematic modeling. Then, the overall transformation from the task space and configuration space is introduced as follows:
\begin{equation}
\left\{\begin{aligned}{}
\textbf{\textit{T}}_E^{B} &=\textbf{\textit{T}}_{R E}^{B} \textbf{\textit{T}}_{F B}^{R E} \textbf{\textit{T}}_F^{F B} \textbf{\textit{T}}_E^F \textbf{\textit{R}}_z\left(\theta_{r z}\right) \\
\textbf{\textit{q}}_e &=\textbf{\textit{q}}_e^{\prime}+\textbf{\textit{q}}_e^{\prime} \textbf{\textit{K}}_c
\end{aligned}\right.
\end{equation}
where $\textbf{\textit{T}}_E^{B}$ is the transformation from the base of the robotic arm to the endpoint of the endoscope. $\textbf{\textit{T}}_{F B}^{R E}$ is the transformation from the end-effector of the robotic arm to the base of the endoscope. $\textbf{\textit{R}}_z\left(\theta_{r z}\right)$ describes the rotation around the $z$-axis to compensate for the errors from the manual assembly. 
$\textbf{\textit{q}}_e^{\prime}$ is the robotic flexible endoscope configuration vector before compensation. 
$\textbf{\textit{K}}_c$ is a 4-by-4 diagonal matrix and the diagonal elements are compensation coefficients for $\textbf{\textit{q}}_e^{\prime}$. $\theta_{r z}$ and $\textbf{\textit{K}}_c$ are calibrated in the following experiments.

\subsection{Path Planning for the Flexible Endoscope}

\begin{figure}[!t]
\includegraphics[width=0.49\textwidth]{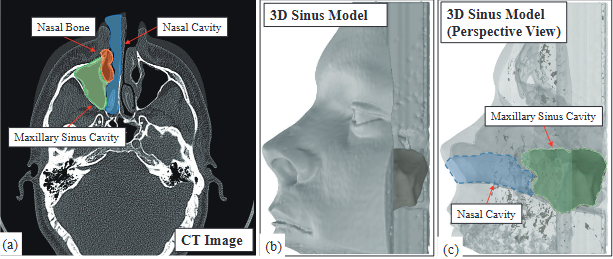}
\centering
\caption{CT image of the sinus and visualization of the 3D sinus. (a) The cross-sectional image of the sinus. (b) 3D sinus model from CT data. (c) Perspective view of the 3D sinus model.}
\label{phantomdata}
\end{figure}

\begin{figure}[!t]
\includegraphics[width=0.49\textwidth]{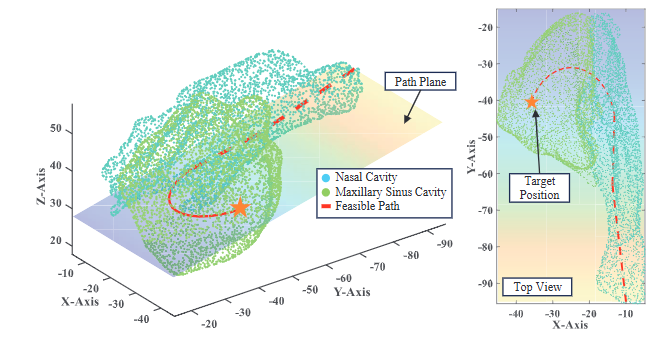}
\centering
\caption{The target position in the maxillary sinus cavity and feasible path obtained by RRT algorithm}
\label{pathplan}
\end{figure}
To achieve an efficient and accurate diagnosis of maxillary sinusitis, the flexible endoscope is required to reach the target position with improved visual coverage. For conducting path planning, the 3D position information of the cloud points is first generated from CT scan data of a realistic 3D sinus phantom \cite{phantom}, as shown in Fig. \ref{phantomdata}. 
The coordinates of these cloud points are unified to the robotic arm coordinate system in subsequent systems registration.
Then, a commonly used rapidly exploring random tree (RRT) algorithm \cite{rrt1, rrt2} is introduced to determine a feasible path for the flexible endoscope to the target position.
The input of the RRT algorithm is a set of cloud points including 1) the cloud points that represent the start position and the target position, 2) the obstacle cloud points ($\textbf{\textit{p}}_{\text {obstacle}}$) that require the flexible endoscope to avoid, representing the environment of the nasal and maxillary sinus cavities.  
%3) and the cloud points ($\textbf{\textit{p}}_{\text {envir}}$) that constitute the environment of the nasal and maxillary sinus cavities. 
The output of the RRT algorithm is a set of target waypoints with 3D position information representing in the phantom coordinate system $\left[\textbf{\textit{X}}_t, \textbf{\textit{Y}}_t, \textbf{\textit{Z}}_t\right]_P$, which form a feasible path. 
In addition, several constraints are considered in the RRT algorithm. 
%Firstly, the endpoint of the endoscope is required to reach the target position, and the configurations of the endoscope and robotic arm cannot exceed their upper and lower limits. 
Firstly, obstacle avoidance is considered. The minimum distance between each waypoint and the obstacle cloud points is required to be bigger than the distance tolerance. 
Secondly, the waypoints are restricted to a certain area shaped by the environment cloud points. That is, to ensure the endpoint of the endoscope is kept in the nasal cavity or maxillary sinus cavity during the diagnosis. 
Finally, in order to keep the endoscope within the natural orifice as much as possible, which will reduce the collisions with the obstacles, the feasible path is required to exist on a plane $\gamma_p$. Three specific cloud points that represent the maxillary sinus ostium, the nostril, and the target position are selected to form the $\gamma_p$.
%$\textbf{\textit{p}}_{\text {target }}$,  $\textbf{\textit{p}}_{\text {no}}$, and $\textbf{\textit{p}}_{\text {so}}$ are on the same plane $\gamma_p$. 
As shown in Fig. \ref{pathplan}, the feasible path on the plane $\gamma_p$ is obtained.

\begin{figure}[!t]
\includegraphics[width=0.47\textwidth]{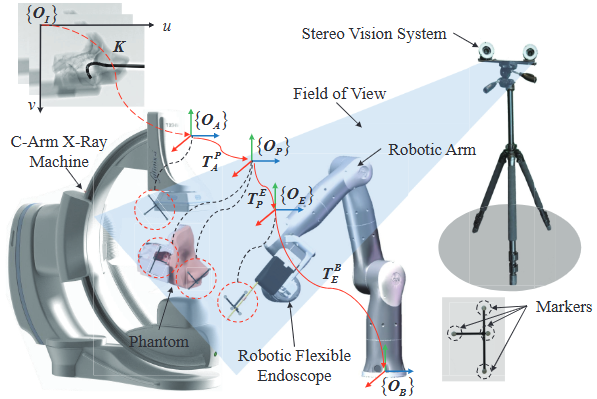}
\centering
\caption{Systems registration among the C-arm X-ray machine coordinate system, phantom coordinate system, and robotic flexible endoscope coordinate system.}
\label{system_regist}
\end{figure}

\subsection{Systems Registration}
%all coordinate systems in the framework are unified on the robotic arm coordinate system. 
Since the feasible path is on the plane $\gamma_p$, which is hard to find and describe in a unified coordinate system for control, registration of the systems is conducted with three steps.

%In order to adjust the orientations of the robotic flexible endoscope, C-arm X-ray machine, and phantom to be on the same plane $\gamma_p$, the Systems registration is conducted with three steps. 
Firstly, as shown in Fig. \ref{system_regist}, several markers are attached to the robotic flexible endoscope, phantom, and C-arm X-ray machine (at least three markers for each to determine a specific plane). 
Secondly, the 3D positions and orientations of the markers are measured by the high-accuracy stereo vision system proposed in our previous work\cite{TIM}. The obtained pose information of the markers constitutes corresponding frames of the C-arm X-ray machine coordinate system $\left\{O_A\right\}$, the phantom coordinate system $\left\{O_P\right\}$, and the endoscope coordinate system $\left\{O_E\right\}$. 
Finally, the orientations of the robotic flexible endoscope, C-arm X-ray machine, and phantom are adjusted, and the registration method proposed in \cite{frameregis} is adopted for systems registration and helps to find the plane $\gamma_p$. 

It is noted that all coordinate systems in the framework are unified on the robotic arm coordinate system. Besides, the $x$-axis and $y$-axis of $\left\{O_P\right\}$ are on $\gamma_p$. Moreover, the transformation from $\left\{O_A\right\}$ to $\left\{O_P\right\}$ is defined as $\textbf{\textit{T}}_A^P$, the transformation from $\left\{O_P\right\}$ to $\left\{O_E\right\}$ is defined as $\textbf{\textit{T}}_P^E$, and the transformation from $\left\{O_E\right\}$ to $\left\{O_B\right\}$ can be obtained from (4). 
Therefore, the transformation from $\left\{O_A\right\}$ to $\left\{O_B\right\}$ is derived as $\textbf{\textit{T}}_{A}^B=\textbf{\textit{T}}_E^{B}\textbf{\textit{T}}_P^E\textbf{\textit{T}}_A^P$, 
which makes $\left\{O_A\right\}$ represents in $\left\{O_B\right\}$. Besides, the $\textbf{\textit{T}}_{P}^B=\textbf{\textit{T}}_E^{B}\textbf{\textit{T}}_P^E$ that represents the transformation from $\left\{O_P\right\}$ to $\left\{O_B\right\}$.

\subsection{Real-Time Image Process}
The real-time image processing system delineates a structured approach for the enhancement of surgical navigation and precision through the endoscope vertex localization in grayscale CT images. This methodology is segmented into three distinct steps, leveraging the capabilities of a multi-modality Large Language Model named LISA \cite{lisa} (large Language Instructed Segmentation Assistant) for advanced image segmentation tasks.

\subsubsection{Step 1: Segmenting the Endoscope with LISA}
The initial step involves the application of LISA for the segmentation of the endoscope from the background within grayscale CT images. LISA combines language processing from large language models with segmentation mask generation. Capable of interpreting complex queries, LISA efficiently generates accurate masks, eliminating the need for extensive datasets of endoscope images and labels for training. The multi-modal foundation increases flexibility and adaptability, enabling segmentation of diverse objects beyond endoscope images through simple text prompts, suggesting broad applicability in future segmentation tasks \cite{modal}. By using a prompt specifically designed for this task, “Please segment the L shape thin and long catheter, going from right to center of the image”, LISA applies its pre-trained model weights to perform the segmentation. This process isolates the endoscope, facilitating further processing and detailed analysis.

\subsubsection{Step 2: Finding the Centerline of the Segmented Endoscope}

Once the endoscope is segmented, the second step involves extracting its centerline. This process, known as skeletonization, is achieved by converting the binary image of the segmented endoscope into a skeletal representation. This transformation is performed using the skimage library's skeletonize function, which reduces the segmented area to its central axis while preserving its topology and general shape. The centerline extraction provides a simplified yet comprehensive representation of the endoscope's structure, essential for subsequent analyses.

\subsubsection{Step 3: Identifying the Endpoint of the Centerline Curve}
The final step in the image processing sequence involves identifying the endpoint of the centerline curve, which serves as the vertex of the endoscope. This is accomplished by analyzing the skeletonized image to locate its endpoints. The process entails identifying all non-zero points representing the skeleton, followed by calculating their distances from a reference point to ascertain the curve's endpoints. A filtering process is then applied to distinguish the points furthest away, under the assumption that these points are located at the extremities of the curve. The algorithm selects the point with the fewest neighboring points as the endpoint, based on the premise that the end of a curve will exhibit fewer connections than other parts. The coordinate of the actual endoscope endpoint in $\left\{O_I\right\}$ is defined as $\left[\textit{u}, \textit{v}\right]^{T}_I$.

\subsection{Optimal Control Scheme}
In order to follow the feasible path on the plane $\gamma_p$, the robotic arm is required to provide axial displacement of the endoscope, which assists the robotic flexible endoscope in navigating in the nasal cavity. 
When approaching the maxillary sinus ostium, the robotic flexible endoscope is simultaneously required to bend through the ostium and access the maxillary sinus cavity. 
Thus, an optimal control scheme is proposed to realize the efficient and accurate coupling motion. 

The input of the control scheme is the actual homogeneous image coordinate vector $\left[\textit{u}, \textit{v},1\right]^{T}_I$, and the mapping between $\left\{O_I\right\}$ and $\left\{O_B\right\}$ is described as follows.
\begin{equation}
\left[\begin{array}{c}
\textit{u} \\
\textit{v} \\
1
\end{array}\right]_I=\textbf{\textit{K}} \textbf{\textit{T}}_{A}^B \left[\begin{array}{c}
\textit{X}_a \\
\textit{Y}_a \\
\textit{Z}_a \\
1
\end{array}\right]_B
\end{equation}
where \textbf{\textit{K}} is a 3-by-4 coefficient matrix representing the mapping between $\left\{O_A\right\}$ and $\left\{O_P\right\}$. Due to systems registration, the plane equation of $\gamma_p$ is already known, which means that if we determine the position information on the $x$-axis and $y$-axis of a point on $\gamma_p$, the position information on the $z$-axis can be calculated correspondingly. 
Thus, each element in \textbf{\textit{K}} is off-line calibrated. 
$\left[\textit{X}_a, \textit{Y}_a, \textit{Z}_a, 1\right]^T_B$ is the homogeneous vector of the actual 3D position information of the endoscope endpoint representing in $\left\{O_B\right\}$. The points of the feasible path in $\left\{O_B\right\}$ can be obtained from:
\begin{equation}
\left[\begin{array}{c}
\textbf{\textit{X}}_t \\
\textbf{\textit{Y}}_t \\
\textbf{\textit{Z}}_t \\
\textbf{1}
\end{array}\right]_B=\textbf{\textit{T}}_{P}^B \left[\begin{array}{c}
\textbf{\textit{X}}_t \\
\textbf{\textit{Y}}_t \\
\textbf{\textit{Z}}_t \\
\textbf{1}
\end{array}\right]_P
\end{equation}

Based on the mapping shown in (5) and (6), the control scheme is described as follows.

\begin{equation}
\begin{aligned}
& \min \quad\quad\quad\quad\quad\quad\quad\quad\boldsymbol{q}^T \boldsymbol{A} \boldsymbol{q} / 2 \\
& \text { s.t. }\left\{\begin{array}{c}
\boldsymbol{J} \dot{\boldsymbol{q}}=\dot{\boldsymbol{x}} \\
\left\|\boldsymbol{p}_{\text {obstacle }}-\boldsymbol{g}^j\left(\boldsymbol{q}_e, \boldsymbol{q}_r\right)\right\|>d_o, j \in[1, \delta] \\
\boldsymbol{q}_{\min }^e \leqslant \boldsymbol{q}_e \leqslant \boldsymbol{q}_{\max }^e \\
\boldsymbol{q}_{\min }^r \leqslant \boldsymbol{q}_r \leqslant \boldsymbol{q}_{\max }^r
\end{array}\right.
\end{aligned}
\end{equation}

where $\textbf{\textit{A}}$ is an 11-by-11 positive definite coefficient matrix, 
$\dot{\textbf{\textit{x}}}=\left(\left[X_a, Y_a, Z_a\right]_B^T-\left[X_t^i, Y_t^i, Z_t^i\right]_B^T\right) / \Delta t$, 
$\Delta t$ is the sample time of the control scheme, 
$\left[X_t^i, Y_t^i, Z_t^i\right]_B^T$ is the $i$-th target waypoint representing in $\left\{O_B\right\}$ and correspond to the actual 3D position information of the endoscope endpoint. 
$\textbf{\textit{J}}$ is a 3-by-11 Jacobian matrix regarding the combination of the Jacobian matrix of the robotic arm and the Jacobian matrix of the flexible endoscope, which are obtained from the kinematics of the robotic arm and our previous work \cite{ref10}, representing the translational velocity contribution to the robotic arm and robotic flexible endoscope. 
%$\dot{\textbf{\textit{q}}}=\left[\dot{\textbf{\textit{q}}}_e, \dot{\textbf{\textit{q}}}_r\right]^{T}$ is the velocity vector of the robotic arm and robotic flexible endoscope. 
%$\textbf{\textit{p}}_{\text {obstacle}}$ represents the obstacle matrix involves the 3D position information of the obstacles that requires the endoscope to avoid. 
$d_o$ is the distance tolerance for obstacles. 
$\textbf{\textit{g}}^j\left(\textbf{\textit{q}}_e, \textbf{\textit{q}}_r\right)$ represents the position vector of the $j$-th body point of the endoscope based on the kinematics in (4), and $\delta$ is the number of the body points. 
%$\textbf{\textit{f}}\left(\textbf{\textit{q}}_e, \textbf{\textit{q}}_r\right)$ represents the position vector of the endpoint of the endoscope based on the kinematics in (4). 
$\textbf{\textit{q}}^{e}_{min}$ and $\textbf{\textit{q}}^{e}_{max}$ are lower and upper joint limitation vectors of the robotic flexible endoscope, respectively. 
$\textbf{\textit{q}}^{r}_{min}$ and $\textbf{\textit{q}}^{r}_{max}$ are lower and upper joint limitation vectors of the robotic arm, respectively. 

The object of the control scheme is to minimize ${\boldsymbol{q}}$ and the output of the control scheme is the configuration vectors of the robotic arm and the robotic flexible endoscope to determine the next coupling motion. 
In addition, the control scheme realizes a closed loop by adopting the difference between the actual and the target 3D position information as feedback, which leads to a higher accuracy control system.

\section{Experiments and Results}
In this section, several experiments were conducted to verify the feasibility of the CT-guided control framework.
\begin{figure}[!t]
\includegraphics[width=0.47\textwidth]{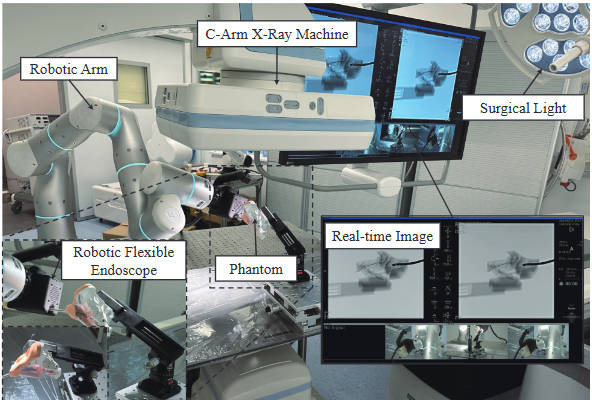}
\centering
\caption{Experimental setup for demonstration of the proposed CT-guided control framework}
\label{fig_setup}
\end{figure}

\begin{figure}[!t]
\includegraphics[width=0.47\textwidth]{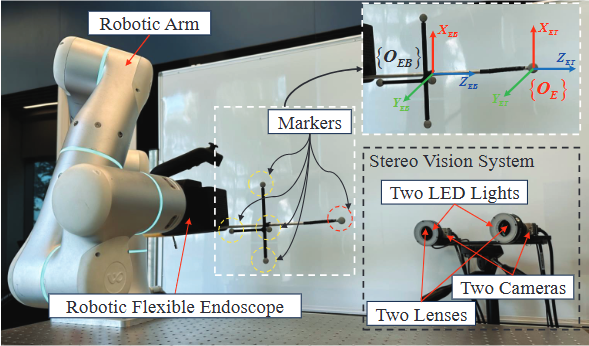}
\centering
\vspace{-0.3cm}
\caption{Experimental Setup for kinematics verification of the robotic flexible endoscope}
\label{setup_kinematics_verification}
\end{figure}

\begin{figure}[!t]
\includegraphics[width=0.4\textwidth]{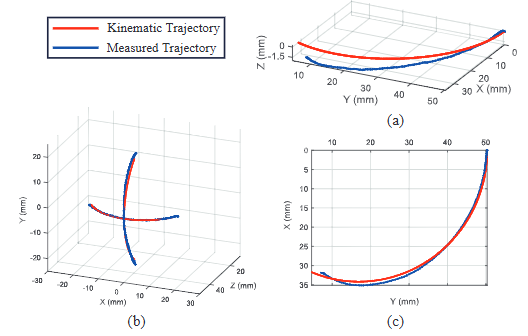}
\centering
\vspace{-0.3cm}
\caption{Experiment results of kinematics verification. (a) 3D kinematic and measured trajectory of the endpoint of the endoscope. (b) 3D kinematic and measured the trajectory of the wrist joints of the endoscope. (c) X-Y plane kinematic and measured trajectory of the endpoint of the endoscope.}
\label{KR}
\end{figure}

\begin{figure}[!t]
\includegraphics[width=0.47\textwidth]{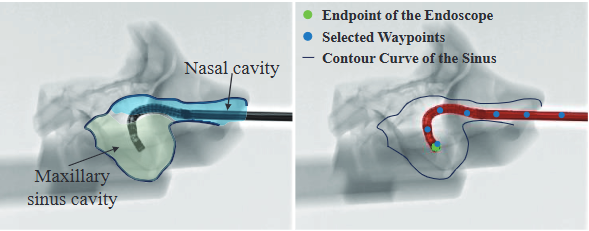}
\centering
\vspace{-0.3cm}
\caption{Real-time original CT image (left) and real-time segmented CT image (right)}
\label{kinematics_verification}
\end{figure}

\begin{figure}[!t]
\includegraphics[width=0.47\textwidth]{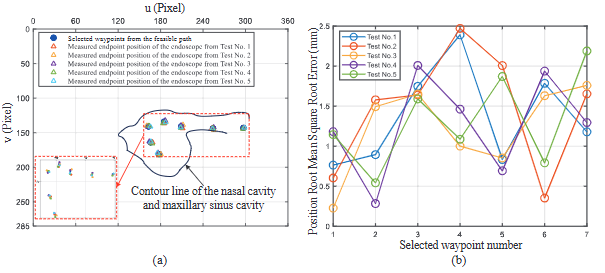}
\centering
\caption{Experiment results of the phantom test. (a) the position of seven selected waypoints from the feasible path and measured endpoint of the endoscope in five tests. (b) root mean square errors of each waypoint in five tests.}
\label{fig1}
\end{figure}

\subsection{Experimental Setup}\label{AA}
The experimental setup is shown in Fig. \ref{fig_setup}. A realistic 3D maxillary sinus phantom (SN-as, PHACON GmbH., Germany\cite{phantom}) was positioned under the C-Arm X-ray machine (Artis Zeego, Siemens, Germany). The robotic flexible endoscope was attached to a commercial 7-DOF robotic arm (Rizon 4, Flexiv, China). In the experiment, the robotic arm was used to control the position and orientation of the robotic flexible endoscope. The robotic flexible endoscope functioned within MATLAB Simulink Real-Time, and Python was employed to control the robotic arm, with communication between the systems facilitated via TCP/IP protocol. Real-time CT videos and videos from several cameras in the hybrid operating room were displayed on the screen.

\subsection{Kinematics Verification}
Several experiments were conducted to verify the kinematics of the robotic flexible endoscope, as shown in Fig. \ref{setup_kinematics_verification}. The robotic flexible endoscope was attached to the end-effector of the robotic arm. In addition, a set of markers was attached to the base of the flexible section, and a marker was attached to the endpoint of the endoscope. The 3D positions of these markers were simultaneously measured by the stereo vision system. 
The robotic flexible endoscope was repeatedly controlled to each joint's limit respectively. At the same time, the 3D position of these markers was measured to calibrate the compensation terms $\theta_{r z}$ and $\textbf{\textit{K}}_c$. The calibrated results were then obtained: $\theta_{r z}$ = 0.0597 rad, the diagonal elements in $\textbf{\textit{K}}_c$ were 0.7255, 0.7255, 0.3435, and 0.7793, respectively. The other experiment was conducted to verify the kinematics with calibrated compensation terms, and the results are shown in Fig. \ref{KR}. From the results, the root mean square position errors of the endoscope (flexible section and rigid wrist) and the wrist joints are 1.11 mm and 0.32 mm, respectively.

\subsection{Phantom Test}
Based on the CT-guided control framework, several experiments were conducted on the phantom. The LISA model was used to find the endoscope endpoint position through CT images, and the endoscope was automatically controlled to follow the feasible path, as shown in Fig. 9. The robotic flexible endoscope successfully followed the feasible path to reach the maxillary sinus cavity. The experiments were repeated five times, resulting in a 100\% success rate of segmentation and identification of flexible endoscope on the CT images. Experimental results (including image process time and recognition success rate) were shown in Table II, revealing an average image process time of 1.16 seconds. The average time (the time for the endoscope to move from the start position to the target position) of the five experiments was 81 seconds.

\begin{table}[!t]
\setlength\tabcolsep{5pt}
\centering
\caption{LISA model process time and recognition success rate}
\begin{tabular}{ccccccc}
\toprule
Experiment number   & 1   & 2   & 3   & 4   & 5  & Mean  \\ \midrule
Image process time (s)        & 1.08 & 1.21 & 1.20 & 1.13 & 1.18 & 1.16 \\
Recognition success rate (\%) & 100 & 100 & 100 & 100 & 100 & 100 \\ \bottomrule
\label{tab_LISA}
\end{tabular}
\end{table}

% 验证精度选择7个点，控制写的时候每个点都follow，follow entire path
% optimized path改成xxx path
To assess control accuracy of the path following, we selected seven waypoints from the feasible path for calculating the error between the actual path and the feasible path. We conducted five experiments and the seven waypoints are shown in Fig. 10(a). Notably, the positioning errors were all within 3 pixels. Through meticulous measurement, we determined that one pixel in the $\left\{O_I\right\}$ to 0.42 mm in the $\left\{O_P\right\}$ , with the average root mean square error for five experiments calculated at 1.48 mm (Fig. 10(b)). The results validate the feasibility of the CT-guided control framework. 

\section*{Conclusion}
In this paper, a novel CT-guided control framework of the robotic flexible endoscope for DMS is proposed. By adopting the proposed CT-guided control framework, the robotic flexible endoscope is able to autonomously follow the feasible path to reach the maxillary sinus cavity efficiently and accurately.
Several experiments are conducted based on the 3D realistic sinus phantom, and the experimental results show that the root mean square position error of the CT-guided robotic flexible endoscope achieves 1.48 mm.
In the future study, the CT-guided robotic flexible endoscope will be adopted for cadaver experiments for further clinical evaluation, and then the proposed CT-guided control framework will be employed for different medical instruments in various surgeries.

\vspace{12pt}

\end{document}